\documentclass[lettersize,journal]{IEEEtran}

\IEEEoverridecommandlockouts
\usepackage{cite}
\usepackage{amsmath,amssymb,amsfonts}
\usepackage{algorithm}
\usepackage{algpseudocode}
\usepackage{graphicx}
\usepackage{textcomp}
\usepackage{xcolor}
\usepackage{subfigure}
\usepackage{tikz}
\usepackage{float}
\usepackage{textcomp}
\usepackage[english]{babel}
\usepackage[nottoc]{tocbibind}
\usepackage{booktabs}
\usepackage{svg}
\usepackage{pgfplots}
\usepackage{setspace}
\usepackage{dsfont}
\usepackage{multicol}
\usepackage{blindtext}
\usepackage{multirow}

\usepackage{adjustbox}
\def\BibTeX{{\rm B\kern-.05em{\sc i\kern-.025em b}\kern-.08em
    T\kern-.1667em\lower.7ex\hbox{E}\kern-.125emX}}

\renewcommand{\tabcolsep}{1pt}

\begin{document}

\title{Enhancing Robotic Arm Activity Recognition with \\ Vision Transformers and Wavelet-Transformed Channel State Information
\thanks{This material is based upon work supported in part by grants ONR N00014-21-1-2431, NSF 2121121, the U.S. Department of Homeland Security under Grant Award Number 22STESE00001-01-00, and by the Army Research Laboratory under Cooperative Agreement Number W911NF-22-2-0001 (all at Northeastern University). The views and conclusions contained in this document are solely those of the authors and should not be interpreted as representing the official policies, either expressed or implied, of the U.S. Department of Homeland Security, the Army Research Office, or the U.S. Government.}
}

\author{\IEEEauthorblockN{Rojin Zandi\IEEEauthorrefmark{1}, Kian Behzad\IEEEauthorrefmark{1}, Elaheh Motamedi\IEEEauthorrefmark{1}, Hojjat Salehinejad\IEEEauthorrefmark{2}\IEEEauthorrefmark{3}, and Milad Siami\IEEEauthorrefmark{1}\\
}
\IEEEauthorblockA{\IEEEauthorrefmark{1}Department of Electrical and Computer Engineering, Northeastern University, Boston, MA, USA}
\IEEEauthorblockA{\IEEEauthorrefmark{2}Kern Center for the Science of Healthcare Delivery, Mayo Clinic, Rochester, MN, USA}
\IEEEauthorblockA{\IEEEauthorrefmark{3}Department of Artificial Intelligence and Informatics, Mayo Clinic, Rochester, MN, USA}
\{zandi.r, behzad.k,motamedi.e\}@northeastern.edu, salehinejad.hojjat@mayo.edu, m.siami@northeastern.edu
}


\maketitle

\begin{abstract}
Vision-based methods are commonly used in robotic arm activity recognition. These approaches typically rely on line-of-sight (LoS) and raise privacy concerns, particularly in smart home applications. Passive Wi-Fi sensing represents a new paradigm for recognizing human and robotic arm activities, utilizing channel state information (CSI) measurements to identify activities in indoor environments. In this paper, a novel machine learning approach based on discrete wavelet transform and vision transformers for robotic arm activity recognition from CSI measurements in indoor settings is proposed. This method outperforms convolutional neural network (CNN) and long short-term memory (LSTM) models in robotic arm activity recognition, particularly when LoS is obstructed by barriers, without relying on external or internal sensors or visual aids. Experiments are conducted using four different data collection scenarios and four different robotic arm activities. Performance results demonstrate that wavelet transform can significantly enhance the accuracy of visual transformer networks in robotic arms activity recognition.
\end{abstract}


\section{Introduction}
\label{sec:intro}
Franka Emika arms are a series of collaborative robotic arms, designed to work alongside humans~\cite{nair2022r3m}. These arms are known for their lightweight and compact design as well as their advanced capabilities in safety and ease of use. Franka Emika arms are utilized across a diverse range of applications in logistics, warehousing, automotive industry, and healthcare due to their versatility, precision, and ability to perform repetitive and complex tasks. 

Activity recognition in robotic arms is the process of identifying and categorizing specific activities or movements being performed by a robotic arm. Activity recognition is essential for enabling robotic arms to comprehend their surroundings, interact proficiently with objects, and autonomously execute tasks \cite{li2021complicated}. To this aim, robotic arms are generally equipped with various sensors such as accelerometers, gyroscopes, joint encoders, force sensors, and vision systems \cite{alatise2020review}. These sensors gather data about the arm's movements, positions, forces exerted, and the surrounding environment. Activity recognition can pose challenges owing to task complexity, environmental variations, and intricacies and noise within sensory data. Vision systems, such as light detection and ranging (LiDAR), typically demand an unobstructed line of sight (LoS), which could be unfeasible in specific surroundings. Privacy also raises concerns, especially in the context of household robots~\cite{li20152d}.

Wireless sensing and detection leverages wireless signals and arrays of sensors to gather diverse types of data. In wireless communications, channel state information (CSI) refers to the information about the current state of a communication channel, such as its quality, fading, interference, and other characteristics \cite{ma2019wifi}. Changes in the environment, including activities performed by robots and humans, can significantly affect CSI. Collected CSI measurements from Wi-Fi signals can be used for human activity recognition (HAR) using various machine learning methods~\cite{zhang2021widar3,salehinejad2022litehar,salehinejad2023joint,8580915,yousefi2017survey, salehinejad2023contrastive,djogo2024fresnel} and hand movement velocity estimation~\cite{hasanzadeh2023hand}. 

Recently, CSI has also been used for robotic arms activity recognition (RAR)~\cite{zandi2023robot,zandi2023robofisense}. A Franka Emika arm was utilized to perform four different activities. During performing the activities, an access point (AP), sniffer, and transmitter collected CSI measurements.
A convolutional neural network (CNN), trained on the collected measurements, was used for robot arm activity recognition. Continuing from the prior work on RAR through Wi-Fi sensing, which employed CNNs, this study advances RAR using vision transformers (ViT) \cite{dosovitskiy2020image} and compares them with a convolutional neural network with long short-term memory (CNN-LSTM) \cite{mutegeki2020cnn} and basic transformer models \cite{vaswani2017attention}. Our ViT-based approach demonstrates remarkable enhancements over the earlier CNN-based approach, showcasing improved recognition accuracy and generalization. 

\begin{figure*}
    \centering
    \includegraphics[width=0.9\textwidth]{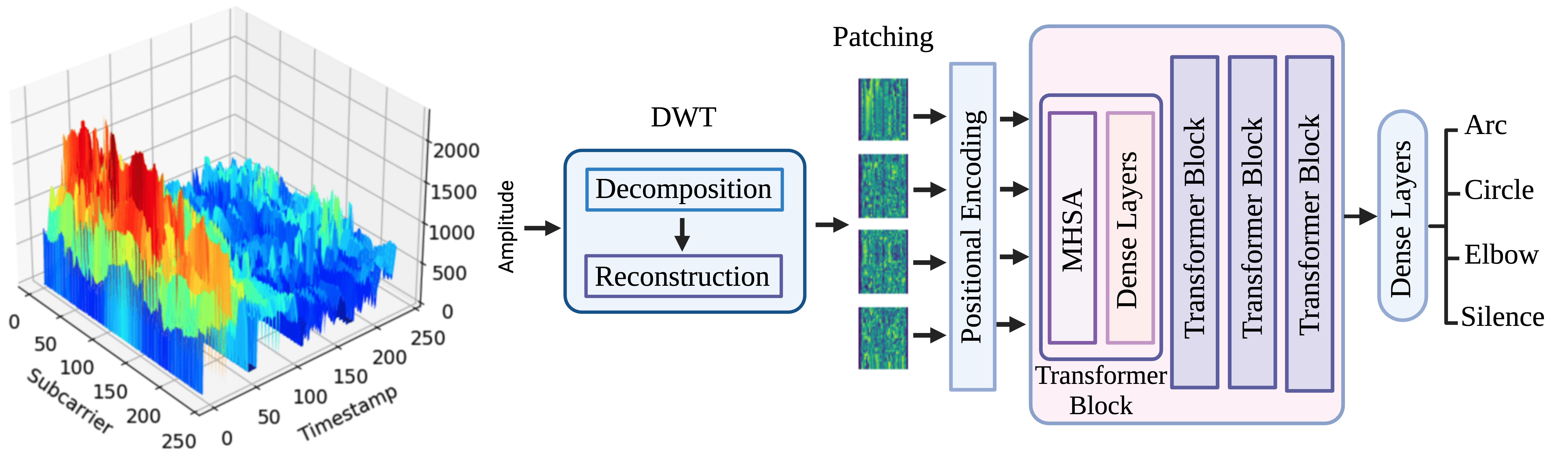}
    \caption{The proposed model for robotic arm activity recognition incorporates discrete wavelet transform (DWT) and multi-head
self-attention (MHSA). The inputs to the model consist of the amplitudes derived from channel state information measurements.}
    \label{fig:process}
\end{figure*}

To further refine our methodology, we have incorporated the discrete wavelet transform (DWT) as a novel noise reduction technique \cite{pasti1999optimization}. Noises such as multipath interference, which creates fading effects due to signals reflecting off surfaces, and co-channel interference from other wireless devices using the same frequency, can obscure the subtle changes in CSI data that are indicative of specific movements or activities. DWT, renowned for its efficacy in signal processing, operates by decomposing a signal into various frequency components, enabling the isolation and reduction of noise while preserving crucial signal features. This addition is particularly significant in Wi-Fi-based activity recognition, where signal clarity is paramount. By achieving superior results on the same dataset, our findings underscore ViT's potential for RAR tasks and contribute to the evolution of Wi-Fi-based activity recognition techniques.
\vspace{-0.25cm}
\section{Proposed Method}
\label{sec:Method}
In the context of RAR, analyzing CSI data from both spatial and temporal dimensions is critical. Temporal analysis reveals the evolution of wireless signal characteristics over time, essential for tracking robotic movements and deciphering complex gestures \cite{sheng2020deep}, while spatial analysis provides insights into signal strength variations and multipath effects across locations. The CSI data is collected as a complex matrix where generally its amplitude is used for activity recognition~\cite{9977629,zandi2023robot,salehinejad2022litehar}. Let ${\{(\mathbf{X}_1, y_1), \dots, (\mathbf{X}_N, y_N)\}}$ represent a set of $N$ CSI measurements where $\mathbf{X}_n \in \mathbb{R}^{S\times T}$ encapsulates the amplitude of the CSI and $y_n\in\{c_1, c_2, \dots, c_{M}\}$ is the corresponding activity class for $M$ number of possible activity classes. Figure~\ref{fig:process} shows different steps of the proposed method where a CSI measurement $\mathbf{X}_n$ is the input and the target class is $y_n$. For convenience, a sample $\mathbf{X}_n$ is represented as $\mathbf{X}$ unless stated. The pseudocode of the model is presented in Algorithm~\ref{alg} and each stage is detailed as follows.

\subsection{Discrete Wavelet Transform for Noise Reduction}

Let $\Tilde{\mathbf{X}}_0=\mathbf{X}$ represent an initial value, decomposed into a set of approximation coefficients using DWT as 
\begin{align}
    \Tilde{\mathbf{X}}_j[i, k] &= \sum_{n=1}^{T} \Tilde{\mathbf{X}}_{j-1}[i, n] \cdot \phi[n - 2k], 
\label{app_row} 
\end{align}
\vspace{-0.25cm}
and 
\vspace{-0.25cm}
\begin{align}    
    \Tilde{\mathbf{D}}_{j}[i, k] &= \sum_{n=1}^{T} \Tilde{\mathbf{X}}_{j-1}[i, n] \cdot \psi[n - 2k], 
\label{det_horiz}
\end{align}
where $\Tilde{\mathbf{X}}_j[i, k]$ are the row-wise approximation coefficients and $\Tilde{\mathbf{D}}_{j}[i, k]$ are the horizontal detail coefficients at level \( j \) using low-pass filter $\phi(\cdot)$ and high-pass filter $\psi(\cdot)$. A 1D DWT is then applied across each column of the row-transformed coefficients as
\begin{align}
    \Tilde{\mathbf{X}}_j[k, l] &= \sum_{m=1}^{S} \Tilde{\mathbf{X}}_j[m, k] \cdot \phi[m - 2l], \label{app_col} 
\end{align}
\vspace{-0.25cm}
and
\vspace{-0.25cm}
\begin{align}
    \hat{\mathbf{D}}_{j}[k, l] &= \sum_{m=1}^{S} \Tilde{\mathbf{X}}_j[m, k] \cdot \psi[m - 2l], \label{det_vert}
\end{align}
where $\Tilde{\mathbf{X}}_j[k, l]$ are the final approximation coefficients and $\hat{\mathbf{D}}_{j}[k, l]$ are the vertical detail coefficients at level $j$. Additionally, the diagonal detail coefficients $\overline{\mathbf{D}}_{j}[k, l]$ can be computed by applying $\psi(\cdot)$ across both dimensions as
\begin{align}
    \overline{\mathbf{D}}_{j}[k, l] &= \sum_{m=1}^{S} \sum_{n=1}^{T} \Tilde{\mathbf{X}}_{j-1}[m, n] \cdot \psi[m - 2k] \cdot \psi[n - 2l],
\label{det_diag}
\end{align}
where the 2D decomposition results in one approximation matrix $\Tilde{\mathbf{X}}_j$ and three detail matrices \( \Tilde{\mathbf{D}}_{j} \), \( \hat{\mathbf{D}}_{j} \), and \( \overline{\mathbf{D}}_{j} \) at each level \( j \), capturing horizontal, vertical, and diagonal details respectively. After each level of decomposition, the approximation coefficients are updated iteratively for $J$ levels. Then, a denoised CSI matrix is computed by applying the inverse 2D DWT as
\begin{equation}
\begin{split}
    \hat{\mathbf{X}}_j[m, k] = &\sum_{l=1}^{\beta}\Tilde{\mathbf{X}}_j[k, l] \cdot \phi^{-1}[2m - l] \\
    &+ \hat{\mathbf{D}}_{j}[k, l] \cdot \psi^{-1}[2m - l],
    \end{split}
\end{equation}
and
\begin{equation}
\begin{split}
    \overline{\mathbf{D}}_{j}[m, k] = &\sum_{l=1}^{\beta} \overline{\mathbf{D}}_{j}[k, l] \cdot \psi^{-1}[2m - l],
\end{split}
\end{equation}
where $\beta=\frac{T}{2^{j}}$. The inverse DWT on rows using the newly obtained column coefficients is computed as
\begin{align}
\begin{split}\label{UC}
    \hat{\mathbf{X}}_{j-1}[m, n] = &\hat{\mathbf{X}}_{j-1}[m, n] \\
    &+ \sum_{k}^{\gamma} \overline{\mathbf{D}}_{j}[m, k] \cdot \psi^{-1}[2n - k],
\end{split}
\end{align}
where $\gamma=\frac{S}{2^j}$. The inverse filters $\phi^{-1}(\cdot)$ and $\psi^{-1}(\cdot)$ are the reconstruction low-pass and high-pass filters, respectively. A denoised matrix is obtained ffter the final reconstruction level, which is the reconstructed version of the original CSI matrix with reduced noise. 

\begin{algorithm}[h!]
\caption{Denoising CSI Amplitude Measurements}
\begin{algorithmic}[1]
\State Input: CSI amplitude $\mathbf{X} \in \mathbb{R}^{S \times T}$, number of levels $J$
\State Output: Denoised amplitude matrix $\Tilde{\mathbf{X}} \in \mathbb{R}^{S\times T}$
\Procedure{DWTDenoising}{$\mathbf{X}$, $J$}
\State Initialize $\Tilde{\mathbf{X}}_0 = \mathbf{X}$
    \For{$j = 1 \rightarrow J$}
        \For{$n=1 \rightarrow T$ and $m=1 \rightarrow S$}
            \State Compute approx. coefficients using (\ref{app_row}) \& (\ref{app_col})
            \State Compute detail coefficients using (\ref{det_horiz}), (\ref{det_vert}) \& (\ref{det_diag})
        \EndFor
        \State Update $\Tilde{\mathbf{X}}_{j-1} \gets \Tilde{\mathbf{X}}_{j}$
    \EndFor

    \For{$j = J \rightarrow 1$}
        \For{$l=1 \rightarrow \beta$ and $k=1 \rightarrow \gamma$}
            \State Up-sample and convolve using (\ref{UC})
        \EndFor
    \EndFor
    \State Set $\Tilde{\mathbf{X}} \gets \Tilde{\mathbf{X}}_0$
    \State return $\Tilde{\mathbf{X}}$
\EndProcedure
\end{algorithmic}\label{alg}
\end{algorithm}

\subsection{Feature Extraction with Transformers}
Each input sample $\Tilde{\mathbf{X}}$ is divided to $Z$ small patches ${\hat{\mathbf{X}} \in \mathbb{R}^{P\times P}}$, where $Z=\lceil\frac{S T}{P^2}\rceil$ and $\lceil\cdot\rceil$ is the ceiling function. Then, these patches are flattened as $\mathbf{x} \in \mathbb{R}^{P^2}$ and embedded to a vector using the positional encoder. Each sequence of these embedded vectors must pass through transformer blocks which contain a multi-head self-attention (MHSA), represented as $\mathbf{M}$, and a multi-layer perceptron (MLP) network. Flattening and embedding the $Z$ patches, results in input sequence $\mathbf{\Lambda}\in \mathbb{R}^{Z \times D}$ where $D$ is the latent vector size. The MHSA is an extension of self-attention $\mathbf{{S}}$ which runs $L$ self-attentions. First, one must compute the query $\mathbf{Q}$, key $\mathbf{K}$, and value $\mathbf{V}$ as
\begin{equation}
    [\mathbf{Q}, \mathbf{K}, \mathbf{V}]=\bf{\Lambda} \mathbf{U}_{\mathbf{Q},\mathbf{K},\mathbf{V}},
\end{equation}
where $\mathbf{U}_{\mathbf{Q},\mathbf{K},\mathbf{V}} \in \mathbb{R}^{D \times 3D}$ is the set of parameters learned during training the network and $\mathbf{Q}$, $\mathbf{K}$, $\mathbf{V} \in \mathbb{R}^{Z \times D}$.
Then, the attention weights are computed as
\begin{equation}
    \mathbf{A}= \sigma\left(\frac{\bf{Q} \bf{K}^\top}{D^{\frac{1}{2}}}\right),
\end{equation}
where $\sigma(\cdot)$ is the Softmax function and $\mathbf{A} \in \mathbb{R}^{Z \times Z}$. The self-attention $\mathbf{S}$ is computed as
\begin{equation}    
    \bf{S}= \mathbf{A}\mathbf{V}. 
\end{equation}
A collection of $L$ self-attention headers are concatenated and projected as
\begin{equation}\label{mhsa}
    \mathbf{M}= [\mathbf{S}_1 \oplus \mathbf{S}_{2} \oplus \dots \oplus \mathbf{S}_L] \mathbf{U}_{MHSA},
\end{equation}
where $\mathbf{M}\in \mathbb{R}^{Z \times D}$, ${\mathbf{U}_{MHSA} \in \mathbb{R}^{(L\times D) \times D}}$ and $\oplus$ is the concatenation operator. It is computed by applying standard multiplication on concatenated self-attentions and ${\mathbf{U}}_{MHSA}$. Following the normalization and residual connection, the data passes through an MLP network. It processes the output of the MHSA layer-by-layer to extract more complex features and relationships, and then another round of normalization and residual connection, is done on the features.

\begin{figure}[t!]
\centering
\subfigure[Arc]{\includegraphics[width=0.22 \textwidth]{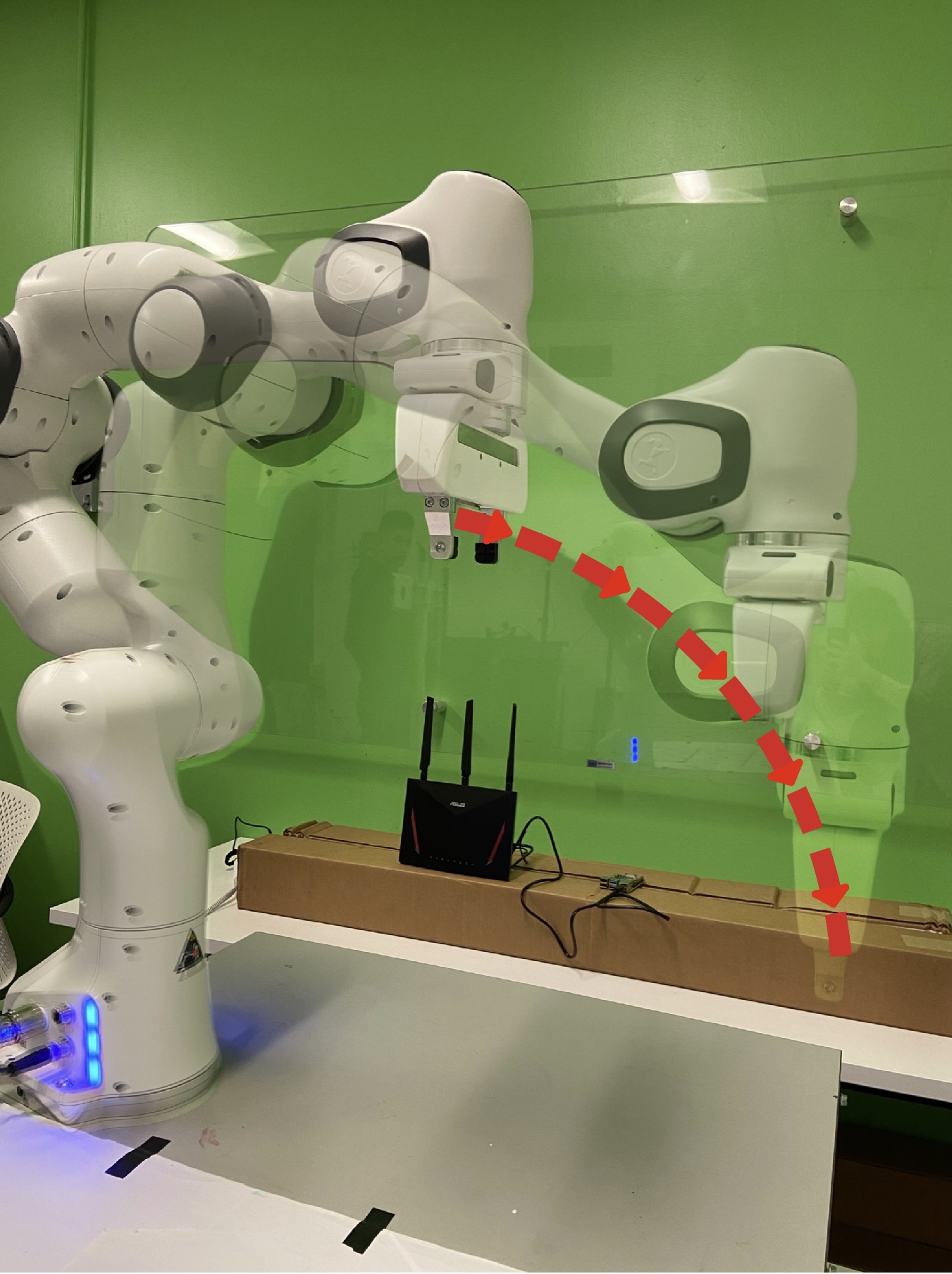}}
\subfigure[Elbow]{\includegraphics[width=0.22 \textwidth]{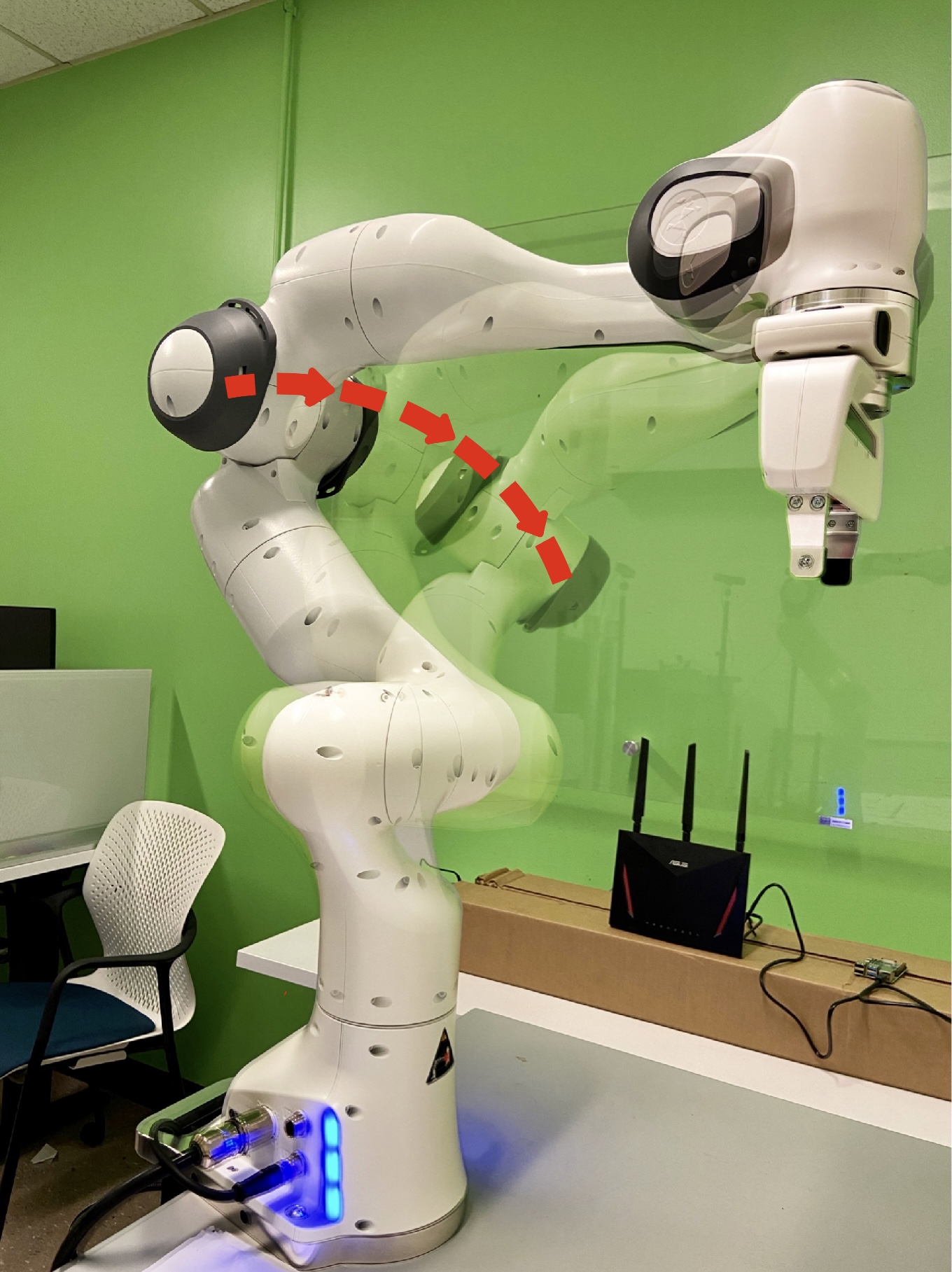}}
\subfigure[Circle]{\includegraphics[width=0.22 \textwidth]{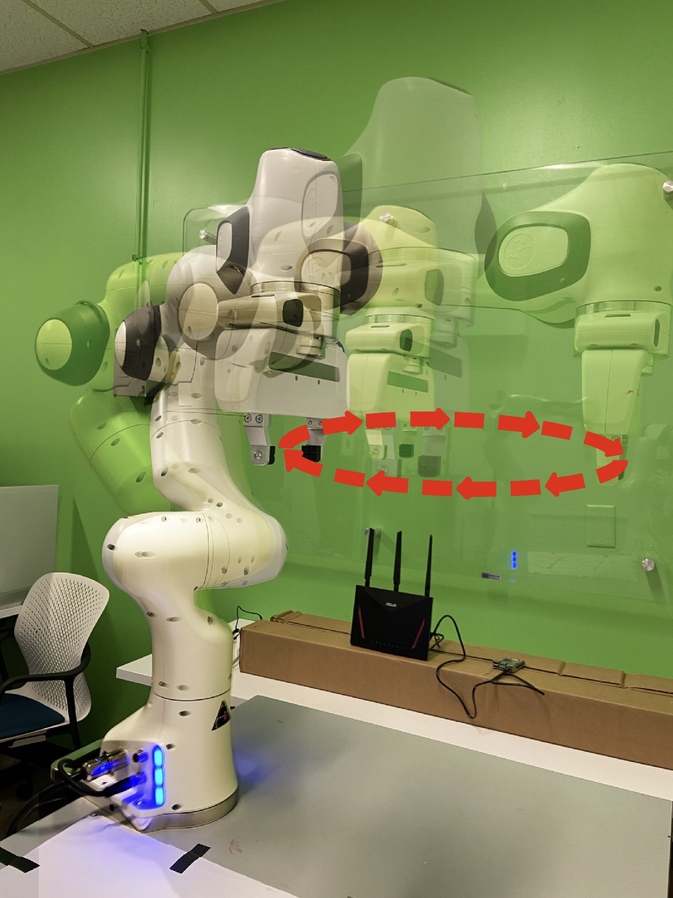}}
\subfigure[Silence]{\includegraphics[width=0.22 \textwidth]{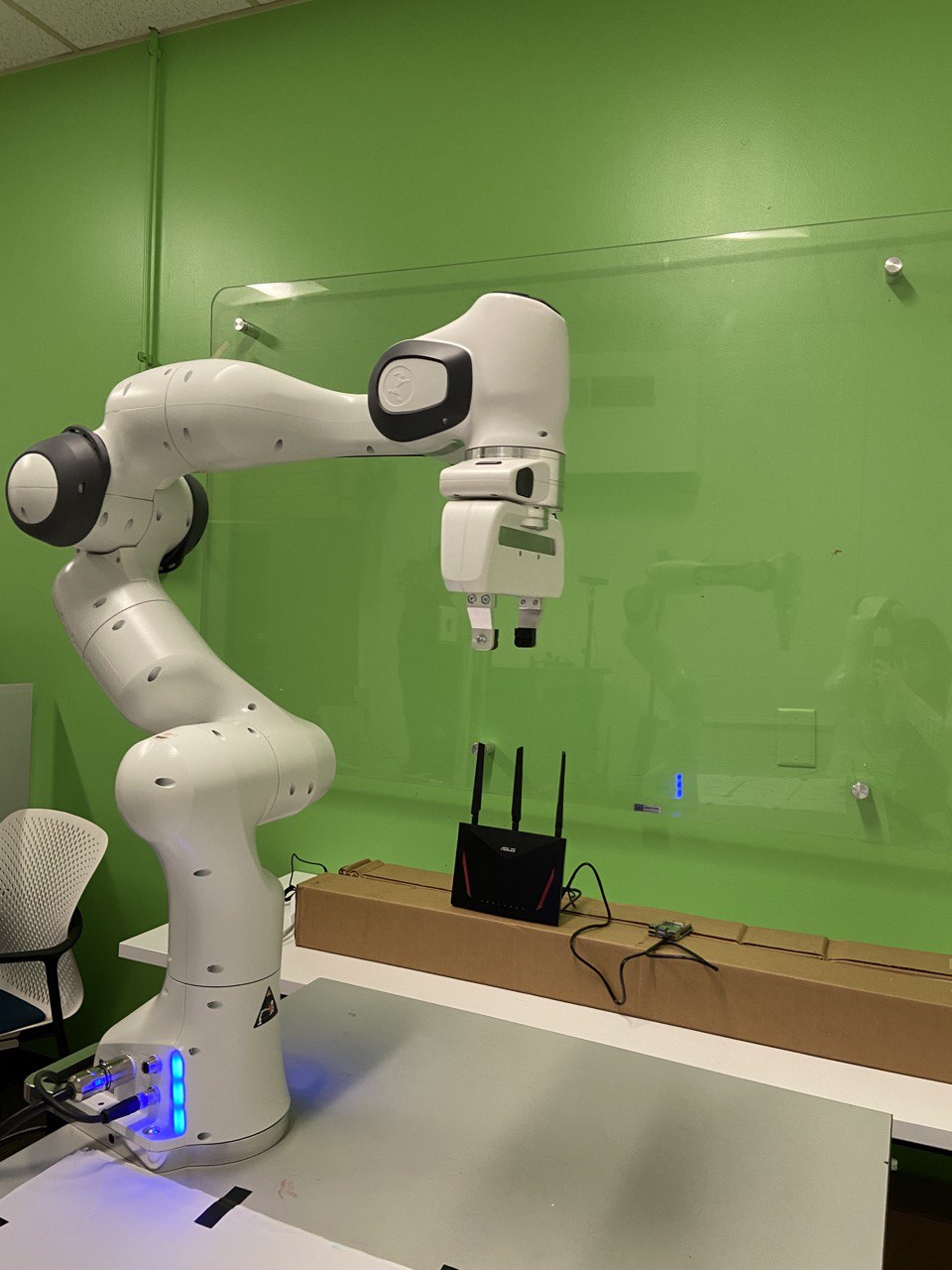}}
\caption{A Franka Emika arm performing four different activities: (a) Arc, (b) Elbow, (c) Circle in the XY-plane, and (d) Silence,~\cite{zandi2023robot}.}
\label{fig:actions}
\end{figure}

\begin{figure}[h]
\centering
\subfigure[Scenario 1]{\includegraphics[width=6cm, height=4.1cm]{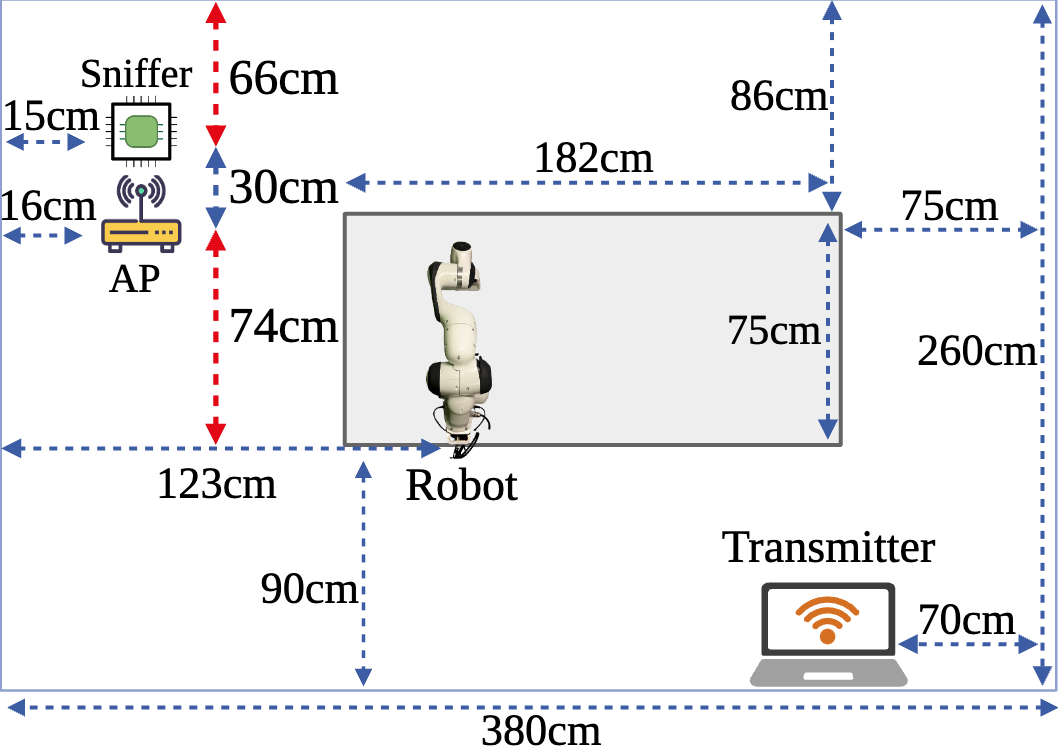}}
\subfigure[Scenario 2]{\includegraphics[width=6cm, height=4.1cm]{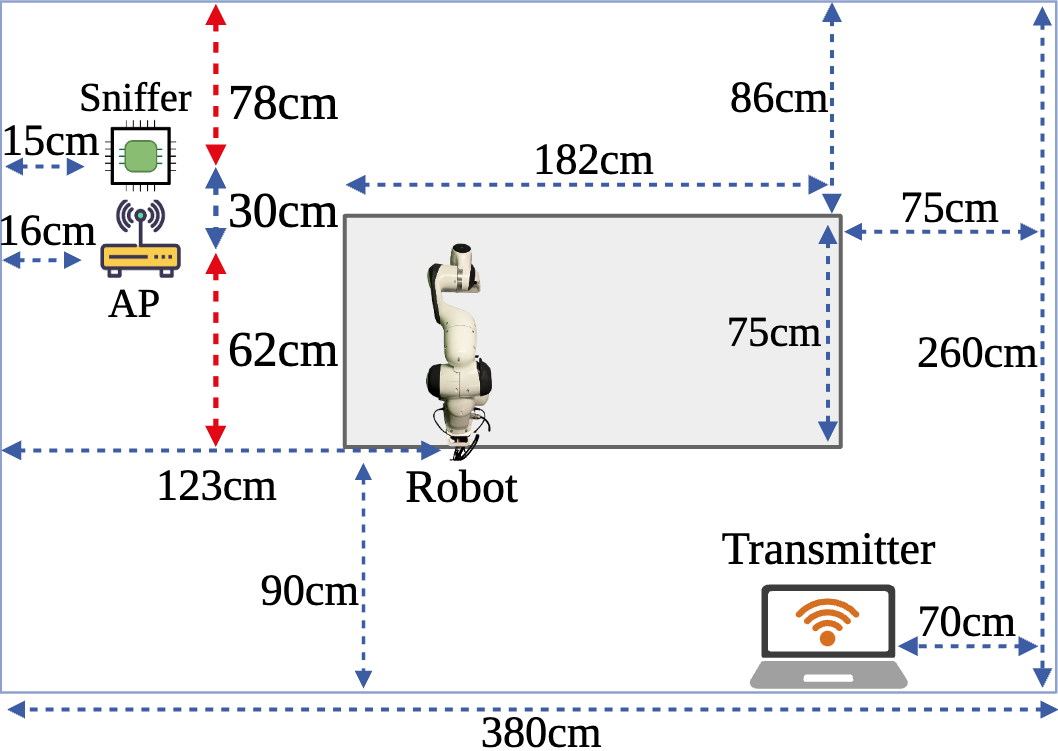}}
\subfigure[Scenario 3]{\includegraphics[width=6cm, height=4.1cm]{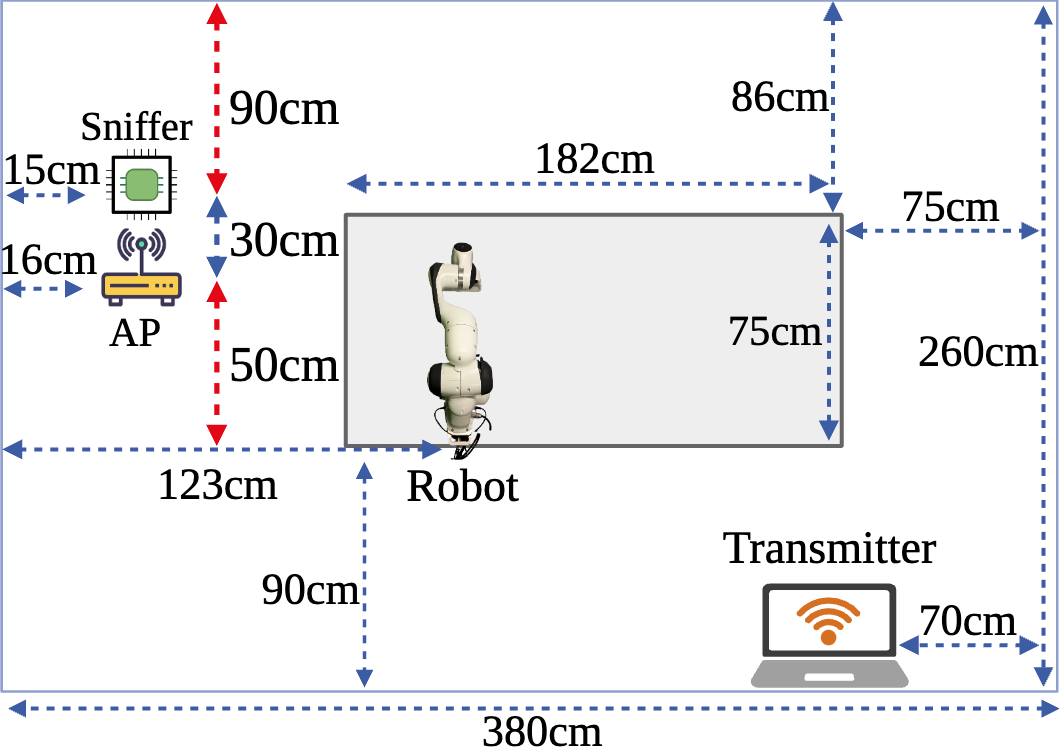}}
\subfigure[Scenario 4]{\includegraphics[width=6cm, height=4.1cm]{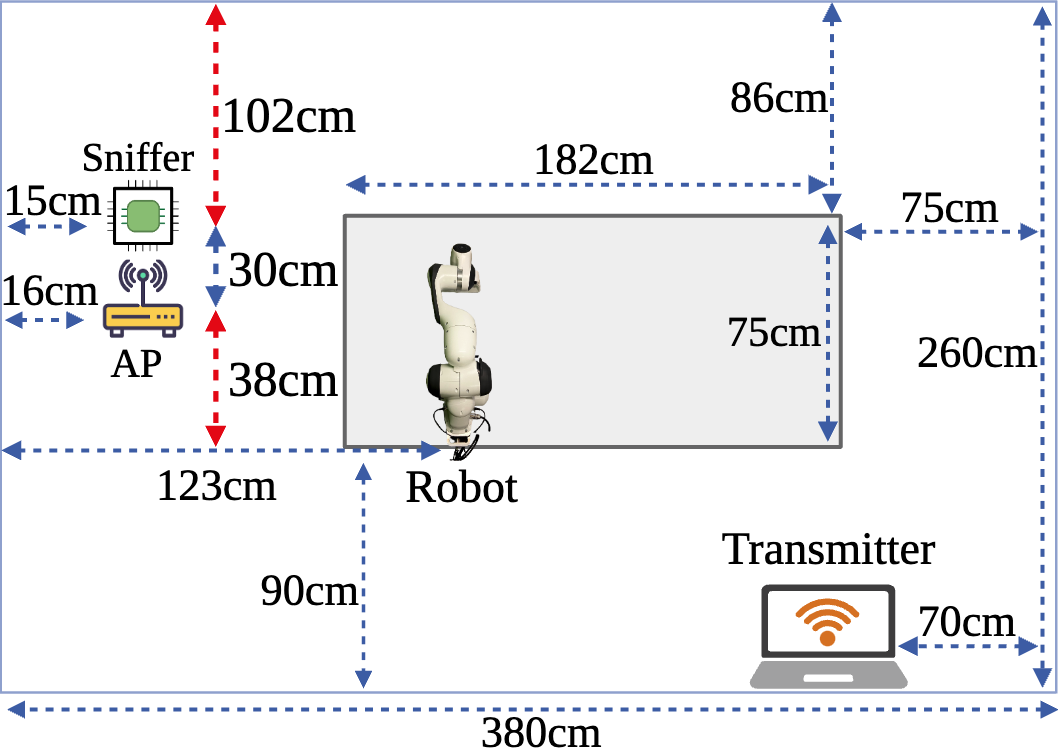}}
\caption{Different data collection scenarios. In each scenario, the red dashed arrows show the change in location of access point and sniffer, which have moved $12$-cm horizontally.}
\label{fig:scanrios}
\end{figure}
\vspace{-0.5cm}
\subsection{Classification}
The output features of the transformer blocks must be fed to the final MLP model to classify the CSI input samples to $M$ classes. The loss function $\mathcal{L}$ of the model is computed for a batch of $N$ samples as
\begin{align}
    \mathcal{L}=-\frac{1}{N} \sum_{n=1}^N \sum_{m=1}^M y_{n, m} \cdot \log \left(\sigma\left(\hat{y}_{n, m}\right)\right),
\end{align}
where $y_{n,m}$ is the true label for sample $\mathbf{X}_n$ and class $m$, and $\hat{y}_{n, m}$ is the is the corresponding logit predicted by the model.
\section{Experiments}
\subsection{Data}
The dataset was acquired from a Franka Emika robotic arm~\cite{zandi2023robot} performing three activities which are (a) Arc, (b) Elbow, (c) Circular motion in the X-Y plane, and (d) No movement (Silence), as presented in Figure~\ref{fig:actions}, based on the floor plans in Figure~\ref{fig:scanrios}. Each activity was repeated for periods exceeding $5$ minutes, capturing data at a frequency of $30$Hz, resulting in the collection of $10,000$ packets by the router. The selected bandwidth was $80$ MHz, utilizing $S=256$ sub-carriers. For every sample, a sequence of $T=300$ consecutive CSI packets was employed. A total of $100$ samples were collected for each activity class. As mentioned in section \ref{sec:Method}, we focus on the amplitude of the CSI data and perform the removal of unused and pilot subcarriers. Consequently, each sample undergoes a reshaping process, resulting in a matrix $\mathbf{X} \in \mathbb{R}^{234\times 300}$, and then DWT with $J=10$ is performed on $\Tilde{\mathbf{X}}$ and finally the denoised signal is ready to be fed to the feature extraction model.

\begin{table}[t!]
\normalsize
    \centering
    \caption{Average and standard deviation of classification metrics of each model, after $10$-fold cross-validation, tested on each scenario in percentage.}
    
\begin{adjustbox}{width=0.4\textwidth}    
\begin{tabular}{|c|c|c|c|c|c|}
\hline
Scenario           & Model       & Accuracy & Precision & Recall & F1-Score\\ \hline
\multirow{5}{*}{1} & CNN         &  90.9$\pm$3.2 & 91.6$\pm$3.9 & 90.9$\pm$3.2 & 90.4$\pm$3.9\\  
                   & CNN-LSTM    & 79.2$\pm$5.3 & 79.4$\pm$6.0 & 79.3$\pm$5.3 & 79.0$\pm$5.6\\ 
                   & Transformer & 69.8$\pm$5.4 & 74.5$\pm$5.2 & 69.8$\pm$5.4 & 70.0$\pm$5.3\\ 
                   & ViT         & 95.0$\pm$1.2 & 95.8$\pm$1.2 & 95.0$\pm$1.2 & 95.0$\pm$1.1\\
                   & ViT-DWT     & \textbf{96.7$\pm$1.2} & \textbf{96.4$\pm$0.7} & \textbf{96.7$\pm$1.2} & \textbf{96.7$\pm$1.2}\\ \hline
\multirow{5}{*}{2} & CNN         & 85.7$\pm$4.1 & 85.6$\pm$4.7 & 85.7$\pm$4.1 & 85.6$\pm$4.6\\ 
                   & CNN-LSTM    & 81.8$\pm$4.7 & 86.5$\pm$4.2 & 81.8$\pm$4.7 & 81.3$\pm$5.1\\ 
                   & Transformer & 63.5$\pm$4.6 & 73.6$\pm$4.3 & 63.5$\pm$4.6 & 63.3$\pm$4.0\\ 
                   & ViT         & 93.7$\pm$1.7 & 94.4$\pm$1.5 & 93.7$\pm$1.7 & 93.7$\pm$1.9\\
                   & ViT-DWT     & \textbf{96.0$\pm$2.2} & \textbf{96.2$\pm$1.3} & \textbf{96.0$\pm$2.2} & \textbf{96.0$\pm$2.1}\\ \hline
\multirow{5}{*}{3} & CNN         & 84.1$\pm$5.4 & 87.0$\pm$4.7 & 84.1$\pm$5.4 & 84.4$\pm$5.0\\ 
                   & CNN-LSTM    & 82.4$\pm$2.7 & 82.8$\pm$1.9 & 82.4$\pm$2.7 & 82.5$\pm$1.5\\ 
                   & Transformer & 70.4$\pm$3.4 & 79.1$\pm$2.8 & 70.4$\pm$3.4 & 70.0$\pm$3.0\\ 
                   & ViT         & 96.9$\pm$1.2 & 97.2$\pm$1.2 & 96.9$\pm$1.2 & 96.9$\pm$1.1\\ 
                   & ViT-DWT     & \textbf{97.4$\pm$1.4} & \textbf{97.7$\pm$2.3} & \textbf{97.4$\pm$1.4} & \textbf{97.3$\pm$2.5}\\ \hline
\multirow{5}{*}{4} & CNN         & 88.1$\pm$4.3 & 88.0$\pm$4.3 & 88.1$\pm$4.3 & 87.8$\pm$3.9\\  
                   & CNN-LSTM    & 83.6$\pm$4.0 & 85.3$\pm$3.7 & 83.6$\pm$4.0 & 83.8$\pm$4.1\\ 
                   & Transformer & 65.4$\pm$5.3 & 75.5$\pm$4.9 & 65.4$\pm$5.3 & 62.3$\pm$4.4\\ 
                   & ViT         & 95.6$\pm$2.0 & 96.0$\pm$2.1 & 95.6$\pm$2.0 & 95.6$\pm$1.8\\
                   & ViT-DWT     & \textbf{97.4$\pm$1.4} & \textbf{97.6$\pm$1.2} & \textbf{97.4$\pm$1.4} & \textbf{97.3$\pm$1.4}\\ \hline
\end{tabular}
\end{adjustbox}
    \label{tab:all_models}
\end{table}

\subsection{Training Setup}
The inherent complexity and sequential nature of CSI measurements come with challenges for conventional pattern recognition methods, which led us to investigate alternative models beyond CNN. In the case of the CNN-LSTM model, we incorporated two LSTM layers into the existing CNN structure. In contrast, for training the transformer model, we flattened each CSI matrix and concatenated them into a vector. 

The ViT model, tailored for CSI classification, employs a patch size of $20$ to capture spatial hierarchies effectively. It is notable that the patches must be square and non-overlapping, so we zero-padded each CSI sample to have input samples with dimension $300\times300$, and after patching each sample results in $225$ patches. It undergoes $60$ epochs of training with a batch size of $64$ and early stopping with $10$ epochs patience, a learning rate of $1 \times 10^{-2}$, dropout rate of $0.3$, and weight decay of $2 \times 10^{-3}$. A grid search was conducted using the validation dataset to tune the hyperparameters. A sensitivity analysis is provided in Subsection~\ref{sec:sensitivity}.

\begin{figure*}[t!]
\centering
\subfigure[Scenario 1]{\includegraphics[width=0.23 \textwidth]{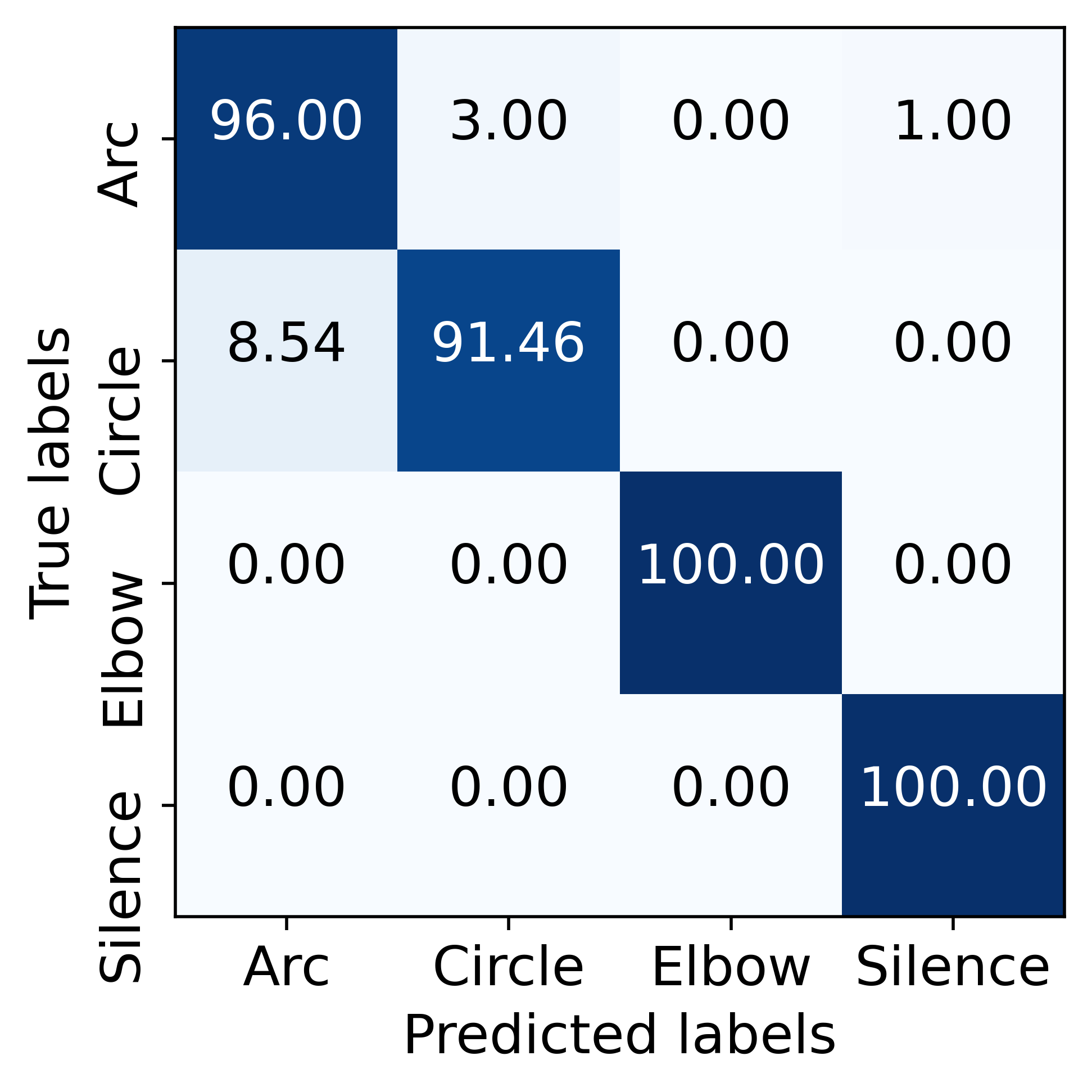}}
\subfigure[Scenario 2]{\includegraphics[width=0.23 \textwidth]{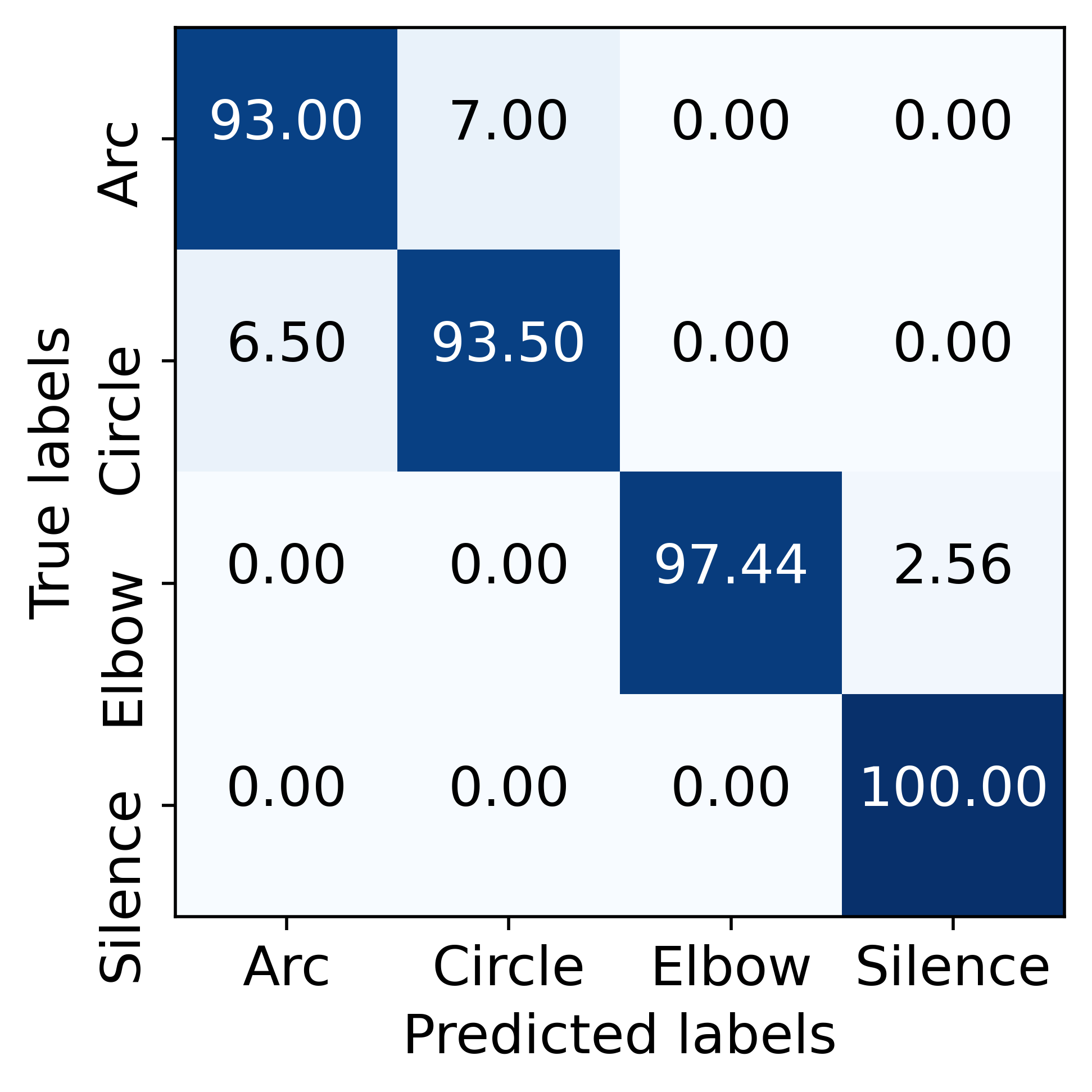}}
\subfigure[Scenario 3]{\includegraphics[width=0.23 \textwidth]{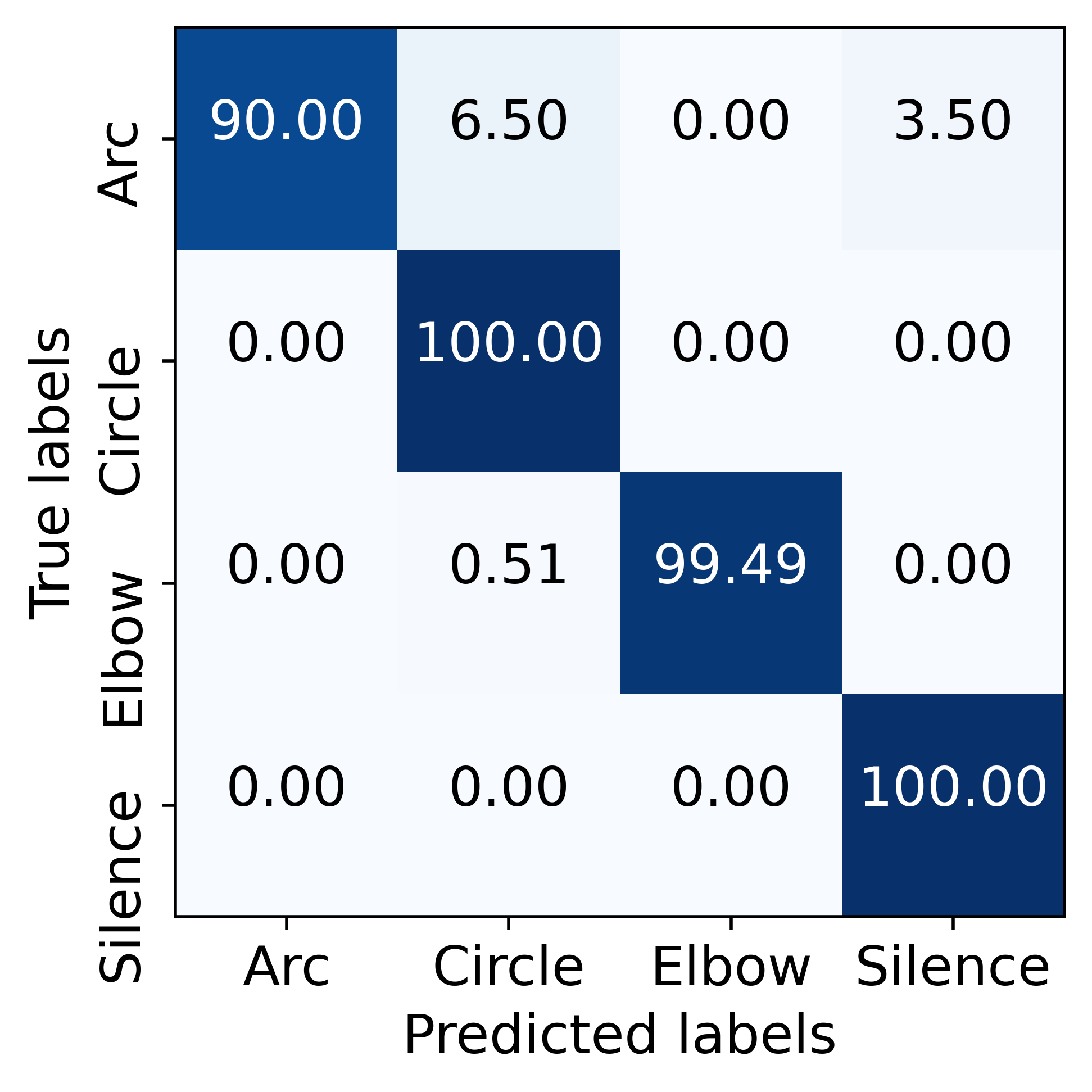}}
\subfigure[Scenario 4]{\includegraphics[width=0.23
 \textwidth]{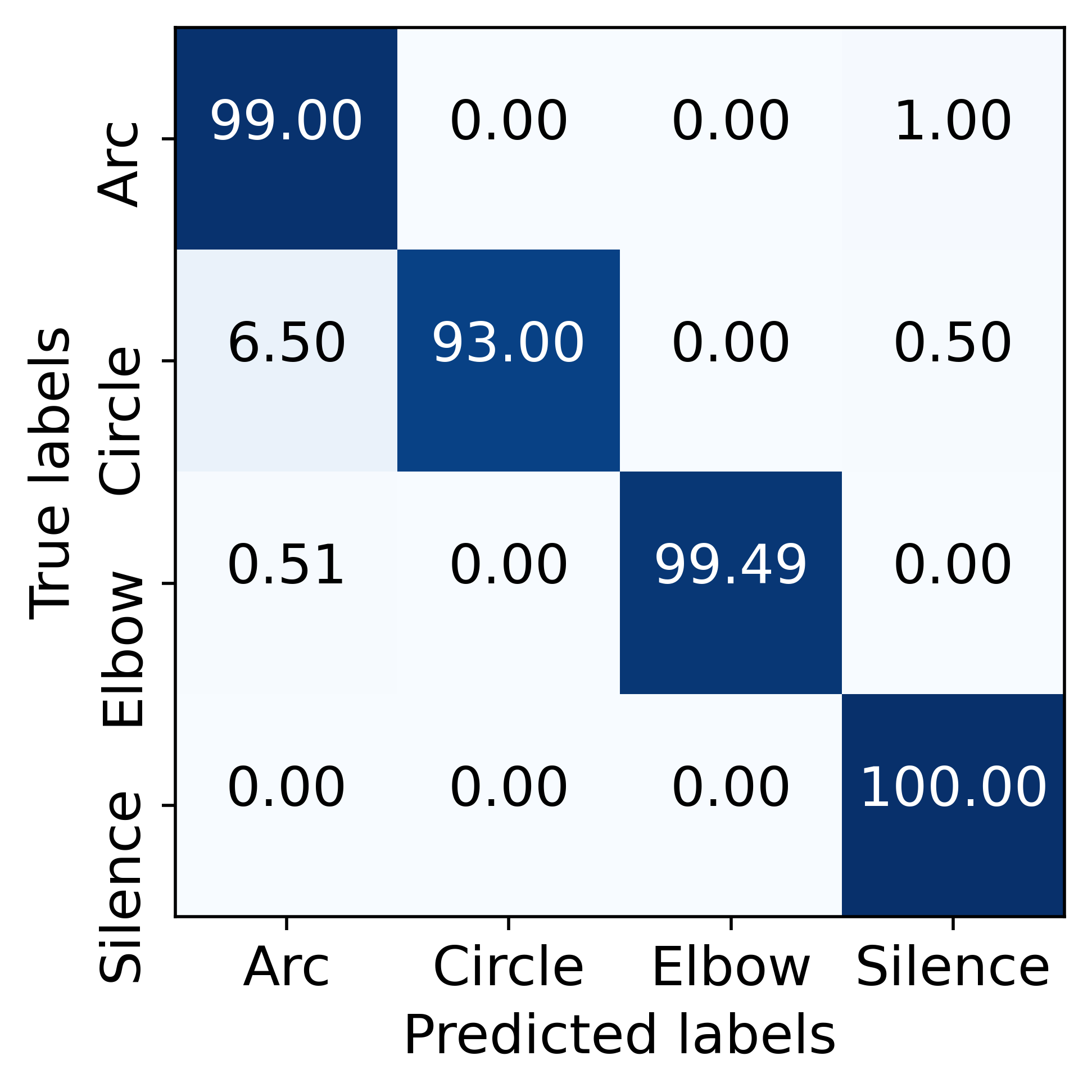}}
\caption{Average confusion matrix values of ViT-DWT, in percentage, after $10$-fold cross-validation for scenarios in Figure~\ref{fig:scanrios}.}
\label{fig:CM-scanrios}
\end{figure*}

\subsection{Results Analysis}
To investigate the impact of altering the sniffer's location, we adopted a leave-one-scenario-out (LOSO) approach, as previously demonstrated in \cite{zandi2023robot}. This strategy involves training our models on three distinct scenarios and testing them on the fourth. In each scenario, both the sniffer and AP were relocated by $12$ cm.


\begin{table}[t!]
\footnotesize
    \centering
\caption{Mean and standard deviation of all leave-one-scenario-out performance evaluations of ViT and ViT-DWT, for each robotic arm activity class, in percentage.}
\begin{adjustbox}{width=0.43\textwidth}    
\begin{tabular}{|c|c|c|c|c|c|}
\hline
Model & Activity & Accuracy & Precision & Recall & F1-Score \\ \hline
\multirow{4}{*}{ViT} & Arc & 90.0$\pm$9.2 & 94.3$\pm$6.5 & 90.0$\pm$9.2 & 91.6$\pm$4.5\\ 
                     & Circle & 95.0$\pm$5.9 & 92.5$\pm$7.9 & 95.0$\pm$5.9 & 93.3$\pm$3.7\\ 
                     & Elbow & 97.4$\pm$4.4 & 100.0$\pm$0.0 & 97.4$\pm$4.4 & 98.6$\pm$2.3 \\ 
                     & Silence & 98.7$\pm$2.2 & 96.7$\pm$5.6 & 98.7$\pm$2.2 & 97.6$\pm$2.8\\ \hline
\multirow{4}{*}{ViT-DWT} & Arc & 94.5$\pm$5.3 & 95.1$\pm$5.3 & 94.5$\pm$5.3 & 94.6$\pm$3.6 \\ 
                     & Circle & 95.9$\pm$5.0 & 94.5$\pm$4.9 & 95.9$\pm$5.0 & 95.1$\pm$3.8  \\
                     & Elbow & 99.1$\pm$2.9 & 100.0$\pm$0.0 & 99.1$\pm$2.9 & 96.0$\pm$4.3  \\
                     & Silence & 100.0$\pm$0.0 & 97.7$\pm$3.2 & 100.0$\pm$0.0 & 98.8$\pm$2.1  \\ \hline
\end{tabular}
\end{adjustbox}
\label{tab:cros-val}
\end{table}

The average and standard deviation of the classification metrics for the CNN, CNN-LSTM, transformer, ViT, and ViT-DWT models are reported in Table\ref{tab:all_models} after $10$-fold cross-validation. Notably, the ViT model exhibits superior performance compared to the CNN model developed in \cite{zandi2023robot}. The ViT model processes an input CSI measurement as a sequence of patches, where the attention mechanism can capture complex temporal and spatial relationships between patches with a higher performance than the CNN and regular transformer networks. 
This aligns with the efficacy of attention-based methods in Wi-Fi sensing, as demonstrated in prior work on HAR~\cite{9745814}. The ViT's superior performance over both CNN and regular transformers underscores the potential of attention mechanisms in this domain.

To have a better understanding of performance of the proposed ViT-DWT model, the average confusion matrix values of each scenario and corresponding classification metrics per activity class are presented in Figure~\ref{fig:CM-scanrios} and Table~\ref{tab:cros-val}. This table presents the average and standard deviation of each class after $10$-fold cross-validation and over all the LOSO experiments for the both ViT and ViT-DWT models. Notably, the ViT-DWT model achieved an average accuracy of over $96.7\%$, with the highest precision recorded for the Elbow activity class. This high precision indicates the model's ability to accurately classify activities without many false positives. Moreover, the model exhibited strong recall rates, emphasizing its capacity to correctly identify instances of each activity class. The F1-Score, which balances precision and recall, further substantiates the model's overall effectiveness.

\subsection{Sensitivity Analysis}
\label{sec:sensitivity}
\begin{figure}[t!]
    \centering
    \includegraphics[width=0.38\textwidth]{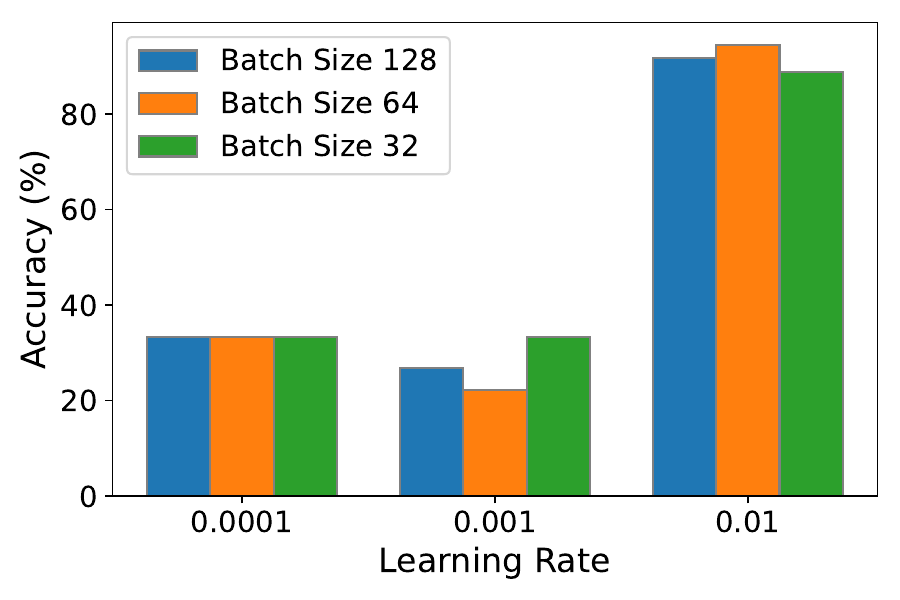}
    \caption{Performance of the ViT-DWT model after grid search for different learning rate and batch size values.}
    \label{fig:hyperparam}
\end{figure}

Sensitivity of the proposed model to different learning rate values $\{10^{-4}, 10^{-3}, 10^{-2}\}$ and batch sizes $\{32, 64, 128\}$ is presented in Figure~\ref{fig:hyperparam} with respect to the accuracy of the model. The results show that a learning rate of $10^{-2}$ and batch size of $64$ achieve the best performance. 
With respect to the activation function, by adopting rectified linear unit (ReLU), we enabled the model to capture nonlinear relationships while mitigating vanishing gradient issues. 
These choices collectively enabled our ViT model to achieve exceptional accuracy in classifying different activities. Key to this model's success is the integration of two attention heads per transformer layer. With four transformer layers and $D=256$, this architecture strikes a balance between depth and breadth of feature extraction. 

\section{Conclusion}
\label{sec:conclusion}
Wi-Fi sensing has conclusively demonstrated its efficacy in HAR and RAR, representing a significant advancement in the field. Our investigation transcended traditional CNN approaches by incorporating advanced architectures like CNN-LSTM, transformer networks, and ViT, with ViT emerging as the superior model due to its unparalleled performance. This superiority was further enhanced by denoising CSI measurements with DWT.
Our findings underscore the pivotal role of ongoing innovation in modeling techniques to drive the evolution and refinement of RAR systems. The objective is to achieve greater precision and context awareness in understanding robotic activities, thereby pushing the boundaries of Wi-Fi sensing applications.
\bibliographystyle{IEEEbib}
\bibliography{strings,refs}
\end{document}